%% file: main.tex
\documentclass[sigconf]{acmart}

\renewcommand\footnotetextcopyrightpermission[1]{}

\AtBeginDocument{
  }

\usepackage{algorithm2e}
\usepackage{multirow}
\usepackage{multicol}
\usepackage[table]{xcolor}
\usepackage{booktabs}
\usepackage{amsmath}

\acmConference[KDD '26]{Make sure to enter the correct
  conference title from your rights confirmation email}{}{}

\acmISBN{978-1-4503-XXXX-X/2018/06}

\begin{document}

\title{RADAR: Benchmarking Vision-Language-Action Generalization via Real-World Dynamics, Spatial-Physical Intelligence, and Autonomous Evaluation}

\author{
  Yuhao Chen\textsuperscript{1,*},
  Zhihao Zhan\textsuperscript{1,*},
  Xiaoxin Lin\textsuperscript{1},
  Zijian Song\textsuperscript{1},
  Hao Liu\textsuperscript{1},\\
  Qinhan Lyu\textsuperscript{1},
  Yubo Zu\textsuperscript{1},
  Xiao Chen\textsuperscript{1},
  Zhiyuan Liu\textsuperscript{1},
  Tao Pu\textsuperscript{1,3},\\
  Tianshui Chen\textsuperscript{3,4},
  Keze Wang\textsuperscript{1,2,3},
  Liang Lin\textsuperscript{1,2,3},
  Guangrun Wang\textsuperscript{1,2,3,\textdagger}
  \\
  \small{
    \textbf{Email:} \{chenyh387, zhanzhh6\} (at) mail2.sysu.edu.cn, wanggrun (at) gmail.com
  }
}

\affiliation{
    \institution{
        \textsuperscript{1}Sun Yat-sen University; \textsuperscript{2}Guangdong Key Laboratory of Big Data Analysis and Processing;\\
        \textsuperscript{3}X-Era AI Lab;\textsuperscript{4}Guangdong University of Technology
    }
    \city{}
    \country{}
}

\renewcommand{\shortauthors}{Chen et al.}

\input{sec/0_abstract}



\maketitle

\begingroup
  \renewcommand\thefootnote{\fnsymbol{footnote}} 
  \footnotetext[1]{These two authors contributed equally and share first authorship.}
  \footnotetext[2]{Corresponding author: Guangrun Wang}
\endgroup

\input{sec/1_intro}

\input{sec/2_related_work}

\input{sec/3_methods}

\input{sec/4_experiments}

\bibliographystyle{ACM-Reference-Format}
\bibliography{sample-base}

\clearpage
\appendix
\onecolumn
\input{sec/5_data}

\end{document}

%% file: sec/0_abstract.tex
\begin{abstract}
Vision--Language--Action (VLA) models have achieved remarkable progress in embodied intelligence; however, their evaluation remains largely confined to simulations or highly constrained real-world settings. This mismatch creates a substantial \emph{reality gap}, where strong benchmark performance often masks poor generalization in diverse physical environments. We identify three systemic shortcomings in current benchmarking practices that hinder fair and reliable model comparison. \textbf{First}, existing benchmarks fail to model \textbf{real-world dynamics}, overlooking critical factors such as dynamic object configurations, robot initial states, lighting changes, and sensor noise. 
\textbf{Second}, current protocols neglect \textbf{spatial--physical intelligence}, reducing evaluation to rote manipulation tasks that do not probe geometric reasoning. 
\textbf{Third}, the field lacks scalable \textbf{fully autonomous evaluation}, instead relying on simplistic 2D metrics that miss 3D spatial structure or on human-in-the-loop systems that are costly, biased, and unscalable.

To address these limitations, we introduce \textbf{RADAR} (\textbf{R}eal-world \textbf{A}utonomous \textbf{D}ynamics \textbf{A}nd \textbf{R}easoning), a benchmark designed to systematically evaluate VLA generalization under realistic conditions. RADAR integrates three core components: (1) a principled suite of physical dynamics; (2) dedicated tasks that explicitly test spatial reasoning and physical understanding; and (3) a fully autonomous evaluation pipeline based on 3D metrics, eliminating the need for human supervision.

We apply RADAR to audit multiple state-of-the-art VLA models and uncover severe fragility beneath their apparent competence. Performance drops precipitously under modest physical dynamics, with the expectation of 3D IoU declining from 0.261 to 0.068 under sensor noise. Moreover, models exhibit limited spatial reasoning capability. These findings challenge the assumption that high scores on traditional benchmarks imply robust embodied intelligence, and position RADAR as a necessary bench toward reliable and generalizable real-world evaluation of VLA models.
\end{abstract}

%% file: sec/1_intro.tex
\section{Introduction}
\label{sec:introduction}

\begin{figure}[t]
    \centering
    \includegraphics[width=1.0\linewidth]{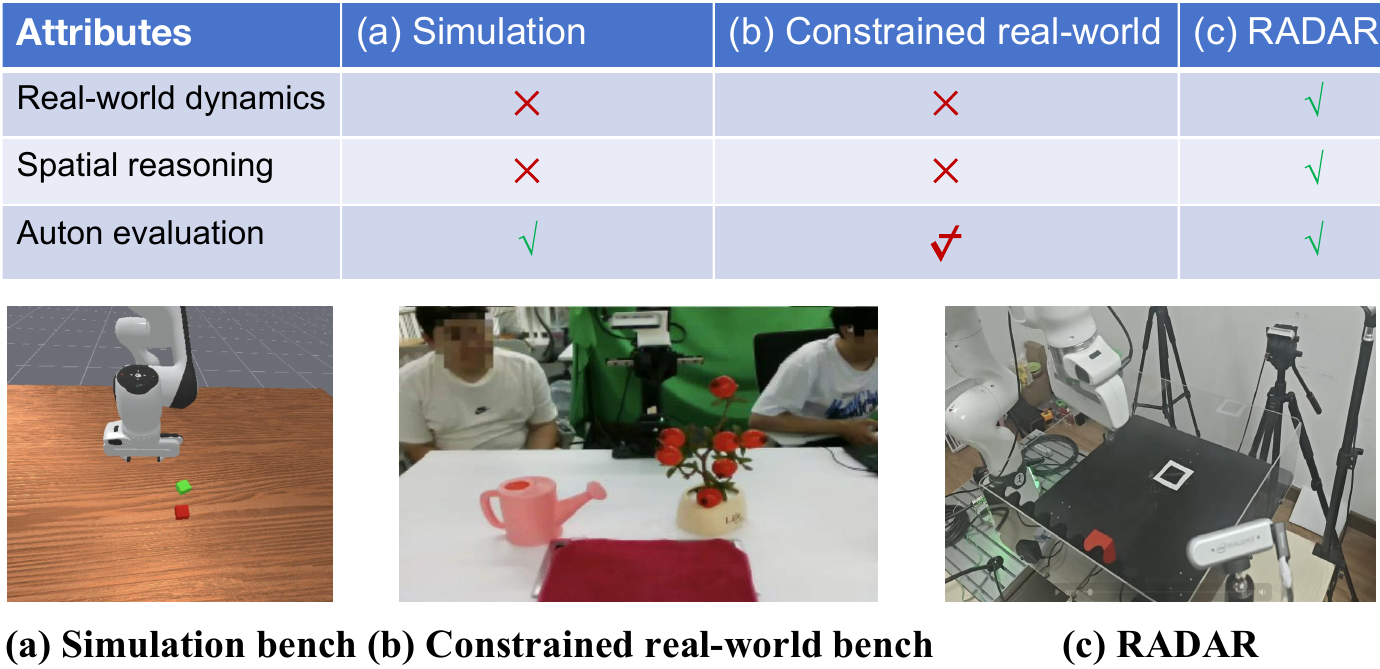}
    \caption{\small{\textbf{A comparison of benchmarking paradigms.} We categorize existing VLA evaluation into Simulation (scalable but lacks physical dynamics) and constrained Real-World setups (realistic but static, lacking spatial consideration, and often relying on manual evaluation). RADAR (Right) bridges these gaps by providing a unified bench that guarantees Real-World Dynamics via systematic dynamics, Spatial Reasoning via geometric tasks, and fully Autonomous Evaluation using high-precision 3D metrics. }}
    \label{fig:cmp}
\end{figure}

The pursuit of general-purpose embodied intelligence has reached a pivotal juncture with the emergence of Vision--Language--Action (VLA) models \citep{zitkovich2023rt,brohan2022rt,black2024pi_0,intelligence2504pi0,zhan2025mathcal,zhan2026stable,li2025vla,song2025physical}. By integrating the semantic reasoning capabilities of large language models with visual perception and robotic control, VLA systems promise agents that can follow natural language instructions in unstructured, real-world environments \citep{chen2026oowm}. Recent studies report impressive results, often achieving near-perfect success rates on established benchmarks. However, a critical disconnect persists: despite strong performance in curated evaluations, these models frequently exhibit fragility and failure when deployed in the wild. We argue that this gap arises not only from a lack of modeling capacity, but also from a systemic failure in current evaluation methodologies.

Existing benchmarks tend to fall into two extremes: \textbf{scalable yet oversimplified simulations} that lack physical dynamics \citep{liu2023libero,robotwin1,chen2025robotwin2,zhang2025vlabench,tao2024maniskill3,zhou2025libero, fei2025libero} (see Fig. \ref{fig:cmp} (a)), or \textbf{constrained real-world setups} that offer limited dynamics and autonomy \citep{zhou2025autoeval,yakefu2025robochallenge} (see Fig. \ref{fig:cmp} (b)). Moreover, both paradigms fail to rigorously evaluate spatial reasoning.

\begin{figure}[t]
    \centering
    \includegraphics[width=1.0\linewidth]{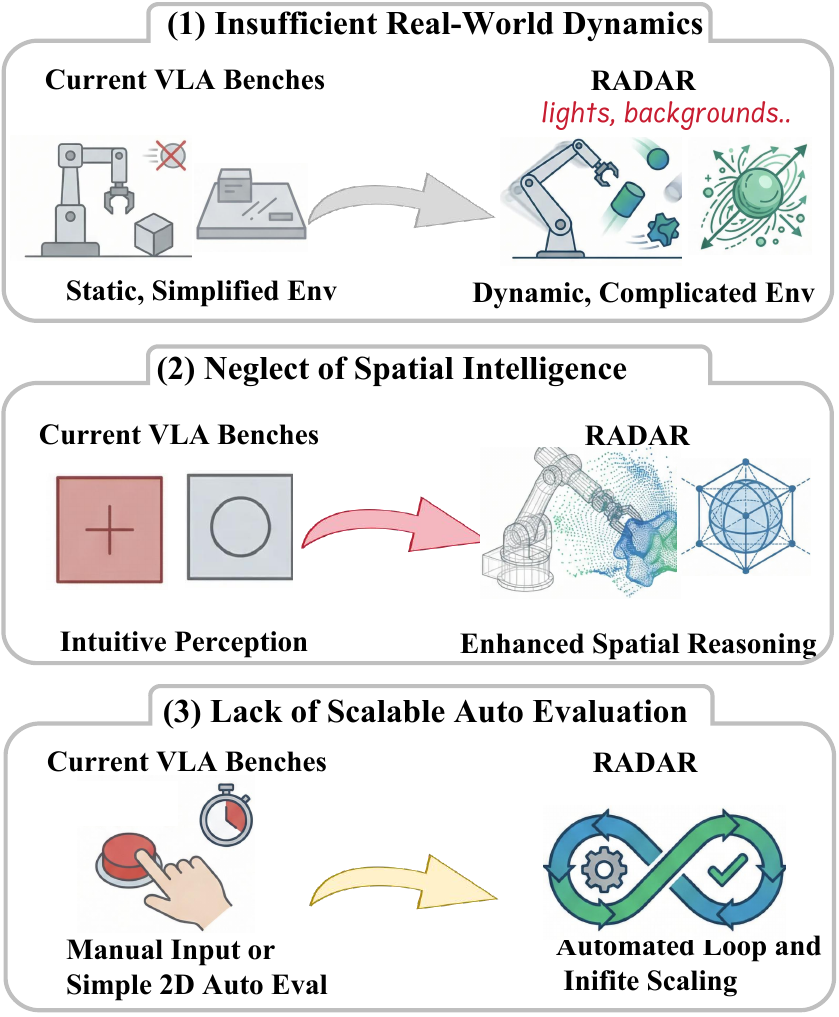}
    \caption{\textbf{A comparison between current VLA benchmarks and the proposed RADAR bench.} Existing benchmarks often rely on (1) static, simplified environments, (2) tasks that require only intuitive perception, and (3) manual or limited 2D evaluation. In contrast, RADAR introduces (1) complex real-world dynamics (e.g., lighting, backgrounds), (2) tasks designed for enhanced spatial-physical reasoning, and (3) a fully automated, 3D-based evaluation loop for infinite scaling. }
    \label{fig:intro}
    \vspace{-11pt}
\end{figure}

This disconnect creates a substantial \emph{reality gap}, where high leaderboard scores mask severe overfitting to static conditions~\citep{zhou2025autoeval,yakefu2025robochallenge}. To address this issue, we focus on \textbf{real-world} evaluation rather than simulation; as illustrated in Fig.~\ref{fig:intro}, we identify three fundamental deficiencies in existing real-world VLA benchmarks that hinder scientific progress and prevent fair model comparison.

\paragraph{Insufficient modeling of real-world dynamics.}
The physical world is inherently stochastic and dynamic, yet many so-called ``real-world'' benchmarks reduce it to a static stage. High-entropy factors that critically affect deployment viability---including dynamic object configurations, robot initial states, lighting variations, background texture noise, and sensor fluctuations---are often ignored or fixed. Without a principled formulation of these dynamics, benchmarks reward memorization of a specific laboratory configuration rather than robust generalization to realistic conditions \cite{li2025vla}.

\paragraph{Neglect of spatial--physical intelligence.}
True embodied intelligence requires more than executing rote pick-and-place behaviors; it demands an understanding of geometry, three-dimensional spatial relationships, and physical constraints. Current evaluation protocols rarely test whether a model genuinely understands a scene or merely exploits spurious 2D correlations. Dedicated tasks that probe spatial reasoning---such as reasoning about geometric, occlusion, and relative positioning---are largely absent, despite being prerequisites for complex manipulation and interaction \citep{song2025siri}.

\paragraph{Lack of scalable autonomous evaluation.}
Evaluation remains a major bottleneck in real-world robotics. Human-in-the-loop protocols (e.g., challenge-style evaluations) are costly, unscalable, and susceptible to observer bias, forcing researchers to wait for results and limiting experimental throughput \citep{yakefu2025robochallenge}. In contrast, existing automated approaches typically rely on 2D object detection or heuristic success criteria, which lack the fidelity required to verify three-dimensional spatial outcomes or nuanced physical interactions \citep{zhou2025autoeval}.

To bridge this gap, we introduce \textbf{RADAR} (\textbf{R}eal-world \textbf{A}utonomous \textbf{D}ynamics \textbf{A}nd \textbf{R}easoning), a benchmark designed to systematically stress-test VLA generalization. RADAR moves beyond ``happy-path'' evaluation by introducing controlled perturbations across four axes: manipulated objects, robot initial states, task instructions, and environmental conditions. Unlike standard benchmarks that focus solely on simple pick-and-place routines, RADAR incorporates dedicated spatial reasoning tasks to rigorously evaluate a model's physical intelligence and geometric understanding. Crucially, the entire system is built upon a fully autonomous evaluation pipeline that employs rigorous 3D metrics, enabling large-scale, reproducible testing without human intervention.

Using RADAR, we conduct a comprehensive audit of state-of-the-art VLA models and uncover striking brittleness beneath their apparent competence. Our analysis reveals that current models are largely over-fitting specialists that collapse under modest physical variation. In particular, performance is hypersensitive to real-world dynamics: sensor noise causes the expectation of 3D IoU to drop from \textbf{0.261} to below \textbf{0.068}. Moreover, even top-performing models struggle with tasks that require genuine 3D spatial reasoning and instruction following, exposing a fundamental limitation in their embodied understanding.

In summary, our contributions are threefold:
\begin{itemize}
    \item We identify and formalize critical limitations of existing VLA benchmarks, particularly in their treatment of environmental dynamics, spatial--physical intelligence, and evaluation autonomy.
    \item We introduce \textbf{RADAR}, a fully autonomous real-world benchmark that integrates systematic environmental dynamics, dedicated spatial reasoning tasks, and 3D evaluation metrics to rigorously assess robustness and generalization.
    \item We provide a rigorous empirical analysis of state-of-the-art VLA models, quantifying their sensitivity to physical and linguistic variations and revealing critical deficiencies in spatial reasoning.
\end{itemize}

%% file: sec/2_related_work.tex
\section{Related Work} \label{sec:related_work}

\subsection{Vision-Language-Action Models}The paradigm of embodied AI has shifted significantly from specialized primitives to general-purpose Vision-Language-Action (VLA) models. Early works demonstrated the efficacy of large-scale imitation learning \citep{brohan2022rt}, while subsequent approaches have focused on integrating the reasoning capabilities of Large Language Models (LLMs) directly into the control loop \citep{zitkovich2023rt, black2024pi_0}. Recent advances have further pushed the boundaries of these models by scaling up training data and model architecture \citep{intelligence2504pi0, li2025vla, zhan2025mathcal}. Despite their impressive zero-shot generalization on semantic tasks, the generalization of these models against physical perturbations and geometric variations remains under-explored. While models like $\pi_0$ \citep{black2024pi_0} and others \citep{zhan2026stable, song2025physical} show promise, their evaluation is often confined to standard ``happy-path'' scenarios, leaving their performance in high-entropy, dynamic environments an open question.

\subsection{Benchmarks in Robotic Manipulation}Benchmarking in robotics has historically bifurcated into two distinct streams: large-scale simulation and constrained real-world setups.

\paragraph{Simulation Benchmarks.} Simulated environments such as LIBERO family (LIBERO \citep{liu2023libero}, LIBERO-Pro \citep{zhou2025libero}, LIBERO-Plus \citep{fei2025liberoplus}), ManiSkill \citep{tao2024maniskill3}, VLABench \citep{zhang2025vlabench}, and RoboTwin \citep{robotwin1, chen2025robotwin2} provide scalable, reproducible platforms for training and evaluation. These benchmarks allow for rapid iteration and precise metric tracking. However, they suffer from a fundamental \emph{reality gap}: simplified physics engines and sterile visual rendering fail to capture the stochasticity of the real world—such as lighting artifacts, sensor noise, and complex friction dynamics. Consequently, policies that excel in simulation often fail to transfer effectively to physical hardware.

\paragraph{Real-World Benchmarks.}
To address the fidelity gap, several real-world benchmarks have been proposed. However, these setups typically prioritize reproducibility at the cost of diversity. Benchmarks like RoboChallenge \citep{yakefu2025robochallenge} and others \citep{zhou2025autoeval} often utilize fixed table-top settings with static lighting and background conditions. Beyond environmental constraints, current protocols frequently neglect spatial--physical intelligence, reducing evaluation to rote manipulation tasks that do not probe geometric reasoning or holistic scene understanding. Furthermore, the field lacks scalable fully autonomous evaluation, instead relying on simplistic 2D metrics that miss 3D spatial structure \citep{zhou2025autoeval} or on human-in-the-loop systems that are inherently costly, biased, and unscalable. \textbf{RADAR} distinguishes itself by bridging these gaps: it introduces a unified framework that guarantees real-world dynamics, enforces spatial reasoning complexity, and enables high-precision autonomous 3D evaluation.

\subsection{Spatial and Physical Intelligence}The integration of Vision-Language Models (VLMs) into robotics has largely focused on semantic reasoning—interpreting high-level instructions like ``pick up the apple.'' However, genuine embodied intelligence requires \emph{spatial-physical intelligence}: the ability to reason about 3D geometry, occlusion, and relative positioning. Recent work has highlighted the deficiency of current VLAs in spatial reasoning \citep{song2025siri}, noting that models often rely on 2D pattern matching rather than 3D scene understanding. While some datasets focus on spatial relations in static images, there is a lack of dynamic benchmarks that require an agent to manipulate objects based on complex geometric constraints (e.g., ``place the block behind the cylinder relative to the gripper''). RADAR addresses this by incorporating dedicated spatial reasoning tasks that cannot be solved via semantic intuition alone.

\subsection{Autonomous Evaluation in Robotics}Evaluating robotic policies in the real world is notoriously difficult. The ``gold standard'' remains human observation, which is unscalable, expensive, and prone to subjective bias. Recent efforts have attempted to automate this process using VLM-based judges \citep{zhou2025autoeval} or 2D object detectors \citep{zhou2025autoeval}. However, VLM judges are susceptible to hallucinations, and 2D detectors lack the depth information necessary to verify 3D success criteria (e.g., checking if an object is hovering above a target). By leveraging a fully autonomous pipeline underpinned by high-precision 3D metrics, RADAR eliminates the human bottleneck while ensuring that evaluation is grounded in physical reality rather than visual approximation.

%% file: sec/3_methods.tex
\section{The proposed RADAR Bench}

We now present the proposed RADAR benchmark. We begin by outlining the overall evaluation pipeline in Section~\ref{sec:pipeline}. We then detail our protocols for modeling real-world dynamics in Section~\ref{sec:dynamics} and describe the design of our spatial intelligence reasoning tasks in Section~\ref{sec:spatial}. Finally, in Section~\ref{sec:eval}, we present our methodology for scalable, autonomous 3D evaluation.

\begin{figure*}[t]
    \centering
    \begin{minipage}[t]{0.49\textwidth}
        \centering
        \includegraphics[width=\linewidth]{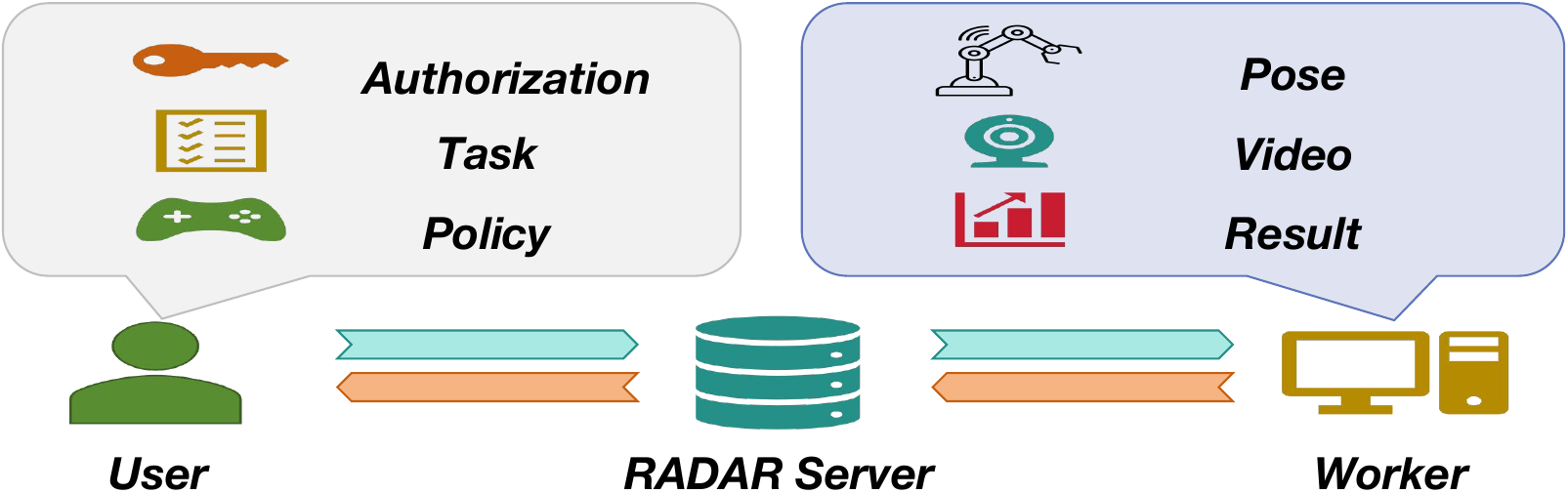}
        \caption{\small{\textbf{The RADAR Pipeline.} RADAR operates as a centralized platform utilizing a client--server--worker architecture. Users submit requests to the central server, which manages authorization, task queuing, and scheduling strategies. The server then dispatches tasks to workers for execution---such as controlling robots or processing video---with the final evaluation results fed back to the user.}}
        \label{fig:insight}
    \end{minipage}
    \hfill
    \begin{minipage}[t]{0.49\textwidth}
        \centering
        \includegraphics[width=\linewidth]{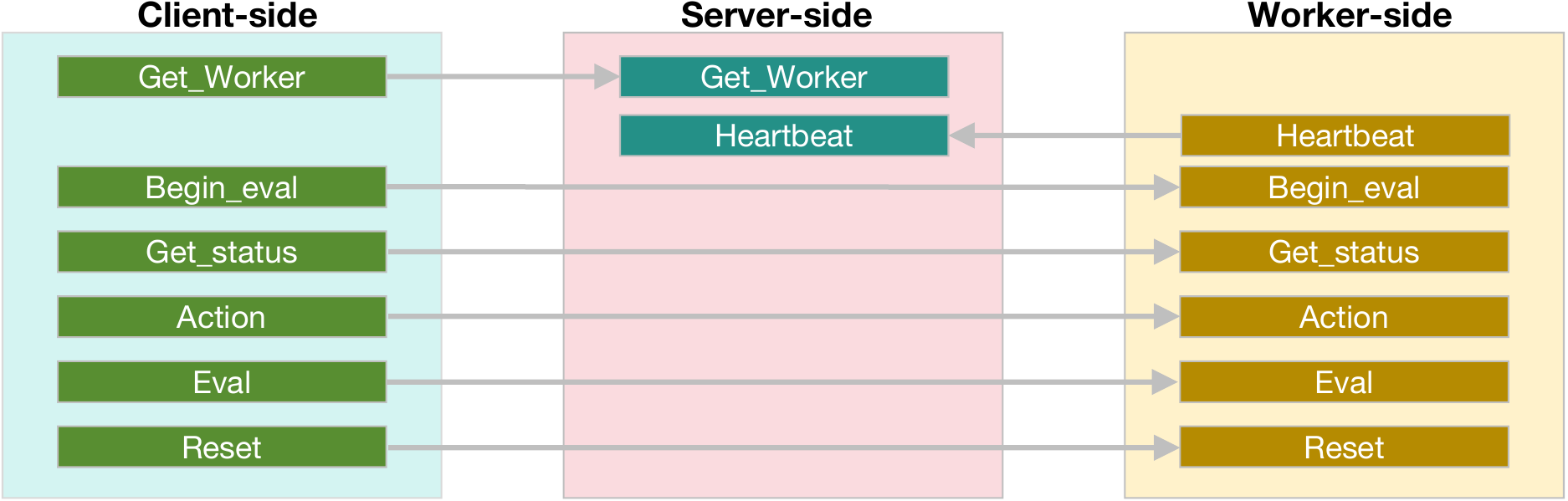}
        \caption{\small{\textbf{RADAR Interface Protocols.} A detailed view of the communication interfaces between the Client, Server, and Worker components. The Client initiates evaluation requests (Begin\_eval, Action, Reset), which are routed to the Worker. The Server coordinates resource allocation via Get\_Worker and monitors system health through Heartbeat signals from workers.}}
        \label{fig:interface}
    \end{minipage}
\end{figure*}

\begin{algorithm}
\SetAlgoLined
\DontPrintSemicolon
\caption{RADAR Evaluation Protocol.}
\label{alg:radar_protocol}

\SetKwInOut{Input}{Input}
\SetKwInOut{Output}{Output}

\Input{Control Policy $\pi$, Task Definition $\mathcal{T}_{task}$, Max Steps $T_{max}$}
\Output{Success Score $S_{eval}$, Trajectory $\tau$}

$\tau \leftarrow \emptyset$ \;
\textbf{Client} sends \texttt{Begin\_eval} request to \textbf{Worker} \;

\For{$t = 0$ \KwTo $T_{max}$}{
    \tcp{Observation Step}
    Receive observation $o_t = \{I_{rgb}, I_{depth}, q_{proprio}\}$ from Worker \;
    
    \tcp{Check Termination}
    \If{IsTerminated($o_t, \mathcal{T}_{task}$)}{
        \textbf{break} \;
    }
    
    \tcp{Policy Inference}
    Compute action $a_t \leftarrow \pi(o_t)$ \;
    
    \tcp{Execution}
    Send $a_t$ to Worker via \texttt{Action} interface \;
    Record tuple $(o_t, a_t)$ to $\tau$ \;
    Wait for execution completion \;
}

\tcp{Automated Scoring}
Request \texttt{Get\_Score} from Server \;
$S_{eval} \leftarrow \text{CalculateIoU}(\text{FinalState}, \mathcal{T}_{target})$ \;

\Return{$S_{eval}$}
\end{algorithm}

\subsection{The RADAR Pipeline}\label{sec:pipeline}

The RADAR benchmark operates on a centralized automation platform designed to decouple policy submission from physical execution. As illustrated in Fig. \ref{fig:insight}, the system adopts a scalable client-server-worker architecture. Users (Clients) act as the entry point, submitting authorization credentials, task specifications, and control policies to a central \textbf{RADAR Server}. This server functions as the system orchestrator; it handles authentication, manages a global task queue, and applies scheduling strategies to dispatch jobs to available \textbf{Worker} nodes.

The specific communication protocols facilitating this workflow are detailed in Fig. \ref{fig:interface}. The interaction begins when the Server assigns a specific Worker to a Client via the \texttt{Get\_Worker} interface. Once connected, the Client communicates directly with the Worker node to minimize latency. The evaluation loop is driven by the Client issuing \texttt{Begin\_eval} requests, followed by high-frequency \texttt{Action} commands. The Worker executes these commands, returning dense geometric data and video logs, while simultaneously sending periodic \texttt{Heartbeat} signals to the Server to confirm system liveliness.

To realize the goal of 24/7 autonomous evaluation within this architecture, the Worker node is implemented as a self-contained robotic cell—the core physical embodiment of the RADAR platform. This integrated hardware system is designed to execute manipulation tasks, reset its own state, and collect rich, multi-perspective sensor data entirely without human intervention. Its architecture is centered around a collaborative robot arm serving as the primary actuator, augmented with a comprehensive sensor suite and an actuated stage to enhance task diversity. The entire setup is orchestrated by a central computing unit, which synchronizes all components to enable seamless, continuous operation. The key components of this platform are detailed below.

\paragraph{Central Computing Unit (The Host PC)} The brain of our system is a high-performance host PC. It runs the core evaluation software, including the task planner, the VLA model inference engine, and the communication middleware (e.g., ROS). This unit is responsible for processing visual data from all cameras, sending control commands to the robot arm and the movable platform, and logging all experimental data—from actuator states to timestamps and performance metrics. Its robust computational power ensures real-time decision-making and synchronization across the entire system.

\paragraph{Robot Arm with Wrist-Mounted Camera}
As the primary manipulation agent, we employ a collaborative robotic arm (e.g., 6-DOF UR5e), equipped with an adaptive gripper (e.g., Robotiq 2F-85). A key component for enabling closed-loop visual manipulation is a camera (e.g., Intel RealSense D435) mounted rigidly on the robot's wrist. This eye-in-hand configuration provides a first-person view from the end-effector, which is crucial for fine-grained manipulation tasks like insertion and precise alignment, offering a perspective that closely matches the agent's action space.

\paragraph{External Stereo Vision System}
To provide a global, third-person view of the workspace and to facilitate tasks requiring broader scene understanding, we integrate two fixed, high-resolution cameras (e.g., Intel RealSense D415). These cameras are strategically positioned at different vantage points around the workspace, forming a stereo vision system. This setup allows for accurate 3D reconstruction of the scene, which is used for initial object localization, tracking agent motion, and providing an objective, global perspective for scoring tasks that require long-horizon reasoning.

\paragraph{Actuated Task Platform}
A critical innovation for achieving full autonomy is our custom-designed actuated platform. This platform, which holds the task-specific objects and environment, is mounted on a mechanical stage (e.g., a linear servo slide or rotating turntable). Its primary function is to enable automatic scene reset. Upon task completion or failure, the host PC can command this platform to move—for instance, by sliding out to a designated "reset position" where a simple mechanism can clear and rearrange objects, or by rotating to present a fresh, pre-configured task scene. This eliminates the single greatest source of manual intervention in real-world testing.

To synthesize the architectural components and interaction protocols described above, we present the unified evaluation workflow in Algorithm \ref{alg:radar_protocol}.

\subsection{Modeling Real-World Dynamics}\label{sec:dynamics}

To rigorously assess the robustness of VLA models, RADAR moves beyond the sterile and static conditions typical of existing benchmarks. In real-world settings, embodied agents must contend with inherent entropy: lighting conditions vary, sensors drift, and human instructions are inconsistent. We operationalize \emph{real-world dynamics} not as a single nuisance factor, but as a principled injection of stochastic perturbations across the physical, visual, and semantic domains during evaluation. We categorize these dynamics into three complementary pillars.

\paragraph{Environmental \& Visual Dynamics.}
The visual appearance of a workspace is rarely constant. We evaluate perceptual robustness by perturbing both photometric and geometric properties of the scene:
\begin{itemize}
    \item \textbf{Illumination Variance.} We introduce randomized lighting conditions to test invariance to shadows and color shifts, varying both light intensity (lux) and color temperature. This forces the agent to rely on geometric cues rather than overfitting to specific pixel intensity distributions.
    \item \textbf{Background \& Distractors.} To challenge segmentation and attention mechanisms, we alter the workspace background, ranging from changes in tabletop texture to the introduction of distractor objects. These distractors share visual attributes (e.g., color or shape) with the target object but are irrelevant to the task.
    \item \textbf{Configurable Layouts.} We reject fixed-position initialization. For each evaluation episode, the initial pose of all objects $\mathbf{T}_{\mathrm{obj}} \in SE(2)$ is randomized within workspace bounds. This encourages the model to reason over relative spatial relationships (e.g., ``left of the mug'') rather than memorizing absolute coordinates.
\end{itemize}

\paragraph{Agent-Centric \& Physical Dynamics.}
A robust policy should be insensitive to specific robot configurations and minor hardware imperfections. We therefore introduce perturbations that affect the agent's physical interface with the environment:
\begin{itemize}
    \item \textbf{Robot State Perturbation.} While standard benchmarks initialize the robot in a fixed ``home'' configuration $\mathbf{q}_{\mathrm{home}}$, we instead sample the initial joint configuration as $\mathbf{q}_{\mathrm{init}} \sim \mathcal{N}(\mathbf{q}_{\mathrm{home}}, \Sigma)$, where $\Sigma$ is a covariance matrix chosen to ensure safe yet diverse starting poses. This prevents policies from overfitting to a single trajectory primitive.
    \item \textbf{Sensor Noise Injection.} To simulate sensor degradation and calibration drift, we inject synthetic noise into the sensory stream, including Gaussian noise on depth maps and small rotational perturbations on extrinsic camera parameters. These perturbations test the resilience of the policy to imperfect state estimation.
\end{itemize}

\paragraph{Semantic Dynamics.}
Real-world instructions are inherently unstructured and variable. To evaluate robustness to linguistic variation, we employ a large language model (LLM) to paraphrase canonical task descriptions into linguistically diverse yet semantically equivalent prompts. For example, the instruction \emph{``Pick up the red block''} may be rewritten as \emph{``Grasp the crimson cube''} or \emph{``Retrieve the scarlet object.''} This forces the VLA model to ground its actions in semantic meaning rather than relying on specific lexical tokens.

\paragraph{Compound Perturbations (``Chaos'' Test).}
Finally, we evaluate agents under \emph{combined dynamics} by sampling jointly from all aforementioned perturbation distributions. In this setting, an agent may encounter dim lighting, a non-canonical initial robot pose, paraphrased instructions, and slight camera miscalibration simultaneously. This ``chaos'' test represents the most challenging evaluation regime, closely mirroring the unpredictability of unstructured real-world environments.

Unlike simulation environments that rely on approximation engines (e.g., MuJoCo, PyBullet) to estimate physical interactions, RADAR validates policies directly in the physical world. Our system implementation ensures that complex dynamic phenomena—such as non-linear friction, deformable object interaction, and contact-rich manipulation—are inherently captured rather than simulated. We implement this through a three-tiered architecture focusing on physical actuation, temporal consistency, and sensory fidelity.

\paragraph{Physical Actuation and Control Interface}
To model realistic contact dynamics, the Worker node operates the robotic manipulator using a high-frequency impedance control loop. Rather than assuming instantaneous state transitions (common in simulation), our implementation accounts for the robot's physical inertia and joint limits.
Let $\mathbf{a}t$ be the action command (e.g., delta-pose $\Delta \mathbf{p} \in \mathbb{R}^6$) sent by the Client. The actual execution on the hardware follows:
\begin{equation}
\mathbf{s}{t+1} = f_{\text{phys}}(\mathbf{s}t, \text{Clip}(\mathbf{a}t, \mathbf{v}{\max})) + \epsilon{env},
\end{equation}
where $f_{\text{phys}}$ represents the true, unknown transition function of the physical world, $\mathbf{v}{\max}$ denotes safety velocity constraints, and $\epsilon{env}$ accounts for unmodeled disturbances (e.g., cable drag or surface friction). By enforcing velocity and acceleration limits, we ensure that the benchmark evaluates policies under realistic actuation constraints.

\paragraph{Temporal Dynamics and Latency Modeling}
Real-world embodied AI often relies on remote inference (cloud robotics) or heavy onboard compute, both of which introduce latency. RADAR explicitly models these temporal dynamics through its asynchronous Client-Worker architecture (refer to Algorithm \ref{alg:radar_protocol}).
The time gap $\Delta t$ between an observation capture $t_{obs}$ and action execution $t_{exec}$ is non-zero and variable:
\begin{equation}
\Delta t = \delta_{net} + \delta_{inf} + \delta_{trans},
\end{equation}
where $\delta_{net}$ is the network round-trip time, $\delta_{\inf}$ is the model inference time, and $\delta_{trans}$ is the hardware transmission latency. This implementation forces evaluated policies to be robust against time-delayed feedback, a critical failure mode in real-world deployment often ignored in zero-latency simulations.

By feeding these raw, noisy signals directly to the agent without ``oracle'' denoising, RADAR evaluates the model's ability to generalize across the imperfections inherent to physical hardware.

\subsection{Spatial Intelligence Reasoning}\label{sec:spatial}

A core objective of RADAR is to evaluate an agent's ability to reason about 3D geometry, spatial relationships, and object affordances. Unlike 2D-centric benchmarks where tasks can often be solved via pixel-matching heuristics, our platform implements spatial intelligence reasoning through three key architectural pillars: multi-view sensory integration, explicit SE(3) action requirements, and volumetric task constraints.

\paragraph{Multi-View Sensory Integration and Occlusion Handling}
Spatial reasoning requires understanding that objects persist and maintain geometry even when occluded. We implement this by positioning cameras at orthogonal viewpoints (e.g., front-down and side-angle). The benchmark provides agents with a stream of calibrated RGB-D data, requiring them to mentally fuse these disjoint perspectives into a coherent scene representation.
To succeed, the agent must resolve ambiguities such as:
\begin{itemize}
\item \textbf{Depth Disambiguation:} Distinguishing between an object's scale and its distance from the camera.
\item \textbf{Occlusion Reasoning:} Predicting the location of a grasp point that is visible in View A but occluded by the robotic arm in View B.
\end{itemize}
We facilitate this by providing precise extrinsic calibration matrices, ${}^{base}T_{cam_i} \in SE(3)$, allowing models to project features from 2D pixel space into the robot's 3D base frame.

\paragraph{High-Precision SE(3) Action Space}
True spatial intelligence manifests in the ability to manipulate objects with 6-Degrees-of-Freedom (6-DoF). We reject simplified action spaces (such as top-down 4-DoF picking) in favor of full rigid-body transformations. The agent must predict an action $\mathbf{a}t = (\Delta \mathbf{p}, \Delta \mathbf{q})$, where $\Delta \mathbf{p} \in \mathbb{R}^3$ represents translation and $\Delta \mathbf{q} \in SO(3)$ represents rotational alignment (represented as quaternions or Euler angles).
This implementation forces the policy to reason about:
\begin{equation}
\text{Alignment Error} = \min{\mathbf{R} \in SO(3)} | \mathbf{R}{obj} - \mathbf{R}{target} |_F,
\end{equation}
where the agent must align the object's principal axes with the target slot (e.g., inserting a rectangular battery into a remote controller), a task impossible without understanding 3D rotational geometry.

\paragraph{Volumetric Task Constraints}
Finally, we implement spatial reasoning requirements through the task definitions themselves. As detailed in the metrics section, success is determined by 3D Volumetric Intersection (IoU) rather than 2D pixel overlap.
This distinguishes ``visual alignment'' from ``physical alignment.'' For example, in a ``mug hanging'' task, a 2D policy might align the mug visually with the hook but fail to place the handle over the hook in depth space. Our evaluation pipeline reconstructs the scene point cloud at every step, ensuring that the agent is penalized for spatial hallucinations (e.g., placing an object behind the target rather than inside it). This strictly enforces that the agent learns a physically valid representation of the workspace.

\begin{figure}[t]
    \centering
    \includegraphics[width=0.9\linewidth]{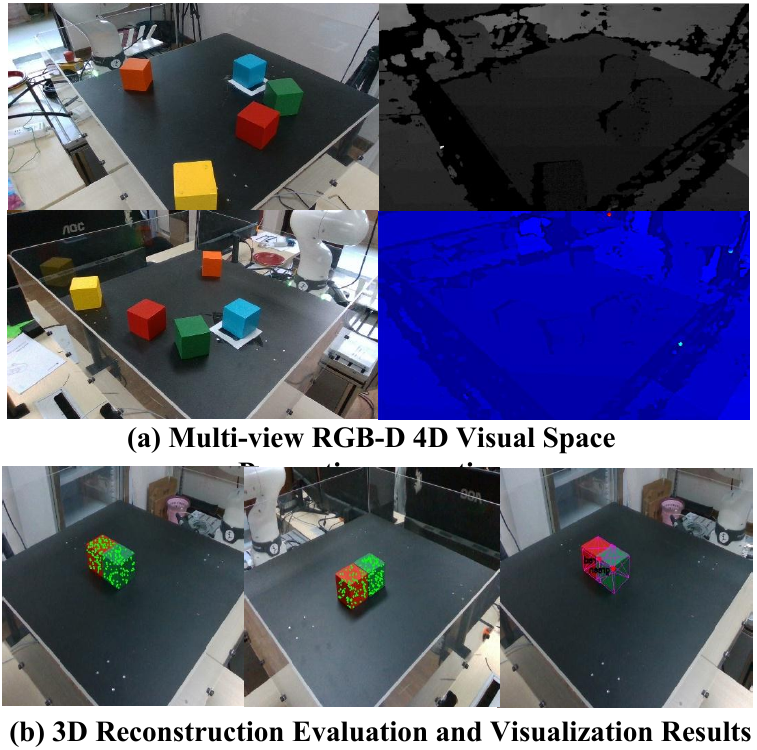}
    \vspace{-11pt}
    \caption{\small{\textbf{Evaluation Metric Computation via 3D Reconstruction.} (a) The system captures Multi-view RGB-D data from the external stereo cameras to create a comprehensive 4D visual representation of the workspace. (b) 3D Reconstruction and Evaluation. The system generates a dense 3D voxel reconstruction of the scene to compare the current object state against the target state.
}}
    \label{fig:eval}
    \vspace{-11pt}
\end{figure}

\subsection{Autonomous Evaluation Metrics}\label{sec:eval}

To ensure standardized and reproducible benchmarking, we define a set of objective and fully automated evaluation metrics for real-world embodied manipulation. As illustrated in Fig. \ref{fig:eval}, our evaluation pipeline leverages multi-view RGB-D sensing to perform dense 3D reconstruction, enabling precise geometric comparisons between the achieved and target states. The metrics are organized into task-level success and geometric accuracy measures, all computed from real-world sensory observations and averaged over $N$ independent trials.

\paragraph{Task Success Rate}
We define task success based on the spatial consistency between the achieved object state and the target state. Specifically, we utilize the 3D Intersection over Union (3D IoU) between the resulting object bounding box and the target bounding box as the primary criterion. As depicted in Fig. \ref{fig:eval}(b), where aligned voxels are visually distinguished from discrepancies, this metric captures the volumetric overlap between the agent's result and the goal. Let $B_{res}$ denote the 3D bounding box of the object after task execution, and $B_{target}$ denote the ground-truth target bounding box. The 3D IoU is calculated as:
\begin{equation}
\mathrm{IoU}_{3D} =
\frac{\mathrm{Volume}(B{res} \cap B_{target})}
{\mathrm{Volume}(B_{res} \cup B_{target})},
\end{equation}
where $\cap$ and $\cup$ represent the intersection and union of the two volumes, respectively. A trial is considered successful if its 3D IoU exceeds a task-specific threshold $\tau$, chosen according to the required tolerance for spatial error. The overall success rate (SR) is defined as:
\begin{equation}
\mathrm{SR} = \frac{1}{N} \sum_{i=1}^{N} \mathbb{1}(\mathrm{IoU}{3D,i} > \tau).
\end{equation}

\paragraph{Translational Error}
To capture fine-grained control accuracy beyond binary success, we quantify the translational error ($e_{trans}$). This metric measures the Euclidean distance between the geometric centroids of the resulting and target bounding boxes:
\begin{equation}
e_{trans} = \frac{1}{N} \sum_{i=1}^{N} \left| \mathbf{c}{res}^{(i)} - \mathbf{c}{target}^{(i)} \right|2,
\end{equation}
where $\mathbf{c}{res}$ and $\mathbf{c}{target} \in \mathbb{R}^3$ are the 3D centers of $B{res}$ and $B_{target}$, respectively.

%% file: sec/4_experiments.tex
\section{Experiments}

\input{tables/result}
\input{tables/training_strategy}
\input{tables/distract}

\begin{figure*}[t]
    \centering
    \includegraphics[width=1.\textwidth]{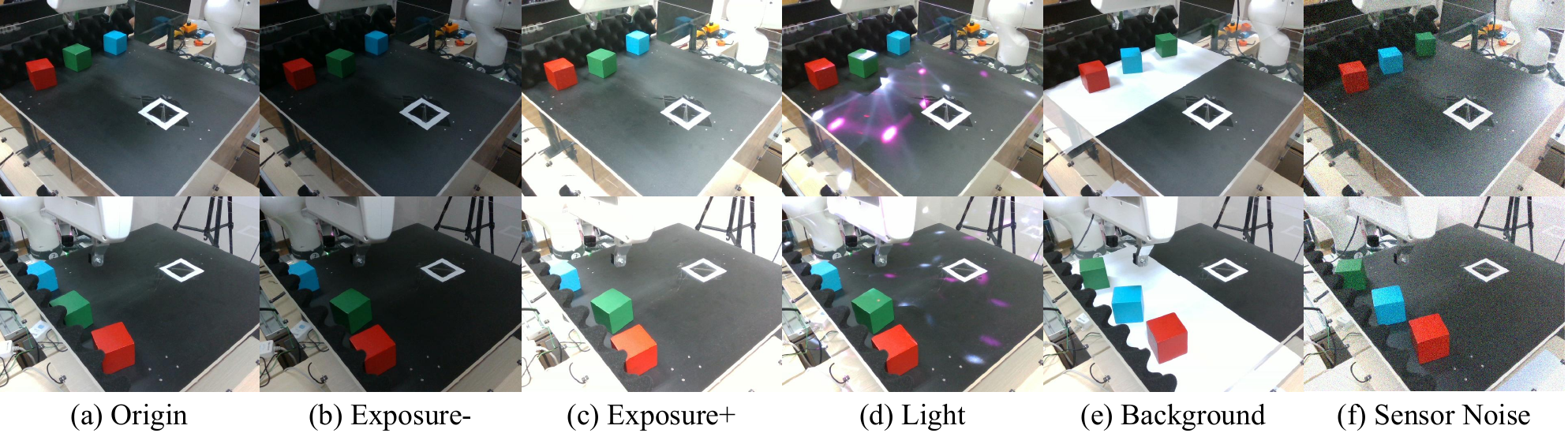}
    \caption{\small{\textbf{Visualization of the different distraction factors.} 
}}
    \label{fig:distract_fig}
    \vspace{-13pt}
\end{figure*}

\subsection{Training Details}

We use $\pi_0$~\citep{black2024pi_0}, $\pi_0$-FAST~\citep{pertsch2025pi0fast}, $\pi_{0.5}$~\citep{intelligence2504pi0}, RDT~\citep{liu2024rdt} and OpenVLA~\citep{kim2024openvla} as baseline models. For all baselines, we initialize from the official pretrained checkpoints released by the respective authors. Each model is fine-tuned on our collected real-robot dataset for approximately 20 hours using a single NVIDIA RTX 6000 GPU. All models are trained for a total of $20{,}000$ optimization steps with a batch size of $32$.

We follow the official training configurations provided in the original implementations~\citep{black2024pi_0,pertsch2025pi0fast,intelligence2504pi0} whenever applicable. The learning rate is scheduled using a cosine decay strategy, with a warm-up period of $1{,}000$ steps, a peak learning rate of $5\times10^{-5}$, and a final learning rate of $5\times10^{-6}$. We employ the AdamW optimizer with a gradient clipping norm of $1.0$.

In addition to the standard settings, we adopt several specific configurations to ensure stable real-robot control on the Franka manipulator during evaluation.
For $\pi_0$, we train the model with an action horizon of $50$ and an action dimensionality of $32$. During evaluation, however, we execute only the first $20$ predicted action steps and retain the first $8$ dimensions (corresponding to 7 joint commands and 1 gripper command). This truncation is applied to improve control stability and reduce compounding errors over long horizons on real hardware.
For $\pi_0$-FAST, we train and evaluate the model using an action horizon of $10$ and an action dimensionality of $8$, following the official design and directly aligning with the Franka control interface.
For $\pi_{0.5}$, we train the model with an action horizon of $10$ and an action dimensionality of $32$. At evaluation time, all $10$ predicted action steps are executed, while only the first $8$ dimensions are used to control the Franka robot, ensuring consistency with the physical action space.
For RDT~\citep{liu2024rdt} and OpenVLA~\citep{kim2024openvla}, we preprocess the data following the official pipeline to obtain the model-required format, and adopt the official training settings for all other configurations.

\subsection{Evaluation Settings}

To comprehensively evaluate the robustness and generalization ability of the models, we construct the test set using four task splits with increasing scene complexity.

\paragraph{Task splits.}
The \textbf{SingleBlock} split contains scenes with only one target object and evaluates basic perception and manipulation capability without relational ambiguity (Fig.~\ref{fig:one}).
The \textbf{TwoBlock} split introduces pairwise object relations, requiring the model to correctly understand relative spatial configurations during task execution (Fig.~\ref{fig:two}).
The \textbf{ThreeBlock} split further increases compositional difficulty by involving multiple objects and more complex interactions (Fig.~\ref{fig:three}).
The \textbf{Spatial} split focuses on relational reasoning between blocks with varied layouts (Fig.~\ref{fig:spatial}). 

\paragraph{Distractor factors.}
For each split, we additionally construct several controlled disturbance conditions to evaluate robustness under appearance variations. 
Specifically, we consider changes in illumination (\textbf{Light}), scene layout (\textbf{Background}), sensor corruption (\textbf{Sensor Noise}), instruction (\textbf{Instruction}), as well as under-exposed and over-exposed imaging conditions (\textbf{Exposure-} and \textbf{Exposure+}). 
All disturbance factors are applied consistently across different splits. (See Fig.~\ref{fig:distract_fig})

\paragraph{Evaluation protocol.}
During evaluation, we directly deploy the trained policy checkpoints. All models are tested under the same data preprocessing pipeline and identical environment configurations to ensure fair comparison. For each split and disturbance condition, multiple evaluation episodes are executed.

\subsection{In-Depth Analysis}

Leveraging the RADAR benchmark, we conduct a comprehensive audit of state-of-the-art Vision–Language–Action (VLA) models. Although these models demonstrate strong performance on standard datasets, our stress-testing reveals notable brittleness under realistic perturbations. In particular, high success rates on static benchmarks often obscure a reliance on shallow heuristics rather than robust embodied understanding (Tabs.~\ref{tab:diff_vla_result}, \ref{tab:mix_result}, \ref{tab:distract_result}).
For instance, under RADAR’s evaluation settings, several representative models—including RDT~\citep{liu2024rdt} and OpenVLA~\citep{kim2024openvla}—fail to complete any tasks successfully.
Based on these observations, we summarize our findings into three primary failure modes:

\paragraph{Hypersensitivity to Physical and Geometric Variation}
Our analysis exposes that current state-of-the-art models operate largely as ``geometric specialists''—overfitting to specific spatial configurations rather than learning generalizable manipulation skills. We observed extreme sensitivity to initial conditions.
This suggests that these models are not ``seeing'' the object in a robust way, but rather memorizing specific pixel-to-action mappings that shatter when the scene geometry is perturbed.

\paragraph{Paradoxical Insensitivity to Language}
Despite being conditioned on natural language instructions, we observe a paradoxical ``blindness'' to semantic nuances. In many trials, models ignored the specific constraints of the prompt (e.g., ``Pick up the \textit{red} block" vs. ``Pick up the \textit{blue} block"), instead defaulting to interacting with the most visually salient object or the object closest to the gripper. This indicates a failure in semantic grounding—the models struggle to adapt their motor policies based on linguistic variation, treating the language input as a weak signal rather than a strict control constraint.

\paragraph{Deficiencies in Genuine Spatial Reasoning}
Finally, our evaluation reveals fundamental limitations in 3D spatial intelligence. While models often succeed at 2D planar tasks (such as pushing an object across a table), they struggle significantly with tasks requiring depth perception and volumetric reasoning (such as precise insertion or stacking). The high failure rate in these tasks highlights a critical gap: current VLA architectures lack a coherent internal 3D world model, relying instead on 2D pattern matching that is insufficient for complex, contact-rich manipulation in the real world.

\paragraph{Extra finding} Finally, we observe that training with the full dataset yields the highest raw action accuracy, which can be observed through the IoU metrics. However, this improvement comes at the cost of increased hallucination, leading to degraded instruction following and reduced semantic fidelity. This trade-off highlights a tension between action imitation performance and reliable language grounding that remains unresolved in current training paradigms and the dataset buildings.

\section{Conclusion}

In this work, we introduced RADAR, a rigorous auditing bench designed to bridge the gap between static evaluation metrics and dynamic real-world robustness. Our evaluation of state-of-the-art VLA models reveals a critical disconnect: despite high performance on traditional benchmarks, current agents exhibit severe fragility when subjected to modest environmental shifts. These findings suggest that current training paradigms prioritize surface-level pattern matching over grounded understanding. By establishing RADAR as a standard for evaluating embodied intelligence, we aim to shift the community’s focus from leaderboard optimization toward the development of agents that are genuinely reliable, generalizable, and robust to the stochastic nature of the real world.

%% file: tables/result.tex
\begin{table*}[t]
\centering
\caption{\textbf{Comparison of different VLA models.} $\mathcal{E}(\text{IoU}_\text{3D})$ is the expectation of $\text{IoU}_\text{3D}$. $\pi_{0.5}$ archive the best result.}
\resizebox{0.95\textwidth}{!}{
\begin{tabular}{c|ccc|ccc|ccc|ccc|cc}
\toprule[1.2pt]
\multirow{2}{*}{Models} & \multicolumn{3}{c}{SingleBlock} & \multicolumn{3}{c}{TwoBlock} & \multicolumn{3}{c}{ThreeBlock} & \multicolumn{3}{c}{Spatial} & \multicolumn{2}{c}{Total} \\ 
& SR & $\text{IoU}_\text{3D}$ & $e_{trans}$ & SR & $\text{IoU}_\text{3D}$ & $e_{trans}$ & SR & $\text{IoU}_\text{3D}$ & $e_{trans}$ & SR & $\text{IoU}_\text{3D}$ & $e_{trans}$ & SR & $\mathcal{E}(\text{IoU}_\text{3D})$   \\ \hline\hline
$\pi_{0}$~\citep{black2024pi_0}           & 2/4 & 0.3284 & 2.07   & 0/10 & \times & \times & 0/10 & \times & \times & 0/8 & \times & \times & 2/32  & 0.0411 \\
$\pi_{0}$-FAST~\citep{pertsch2025pi0fast} & 2/4 & 0.3797 & 1.83   & 0/10 & \times & \times & 0/10 & \times & \times & 0/8 & \times & \times & 2/32  & 0.0475\\
$\pi_{0.5}$~\citep{intelligence2504pi0}   & \textbf{4/4} & \textbf{0.5216} & \textbf{1.08}   & \textbf{4/10} & \textbf{0.4608} & \textbf{1.88}   & \textbf{6/10} & \textbf{0.4816} & \textbf{1.67}   & \textbf{3/8} & \textbf{0.1261} & \textbf{3.8}    & \textbf{17/32} & \textbf{0.2605}  \\
RDT~\citep{liu2024rdt}                    & 0/4 & \times & \times & 0/10 & \times & \times & 0/10 & \times & \times & 0/8 & \times & \times & 0/32 & \times  \\
OpenVLA~\citep{kim2024openvla}            & 0/4 & \times & \times & 0/10 & \times & \times & 0/10 & \times & \times & 0/8 & \times & \times & 0/32 & \times  \\
  \bottomrule[0.5pt]
\end{tabular}
}
\label{tab:diff_vla_result}
\end{table*}

%% file: tables/training_strategy.tex
\begin{table*}[htbp]
\centering
\caption{\textbf{Comparison of baseline models with different training strategy}. Independent means training only with the corresponding data. Mix ALL means training with all the data. Mix Blocks means training without spatial data, which leads to significant failure in spatial task. $\mathcal{E}(\text{IoU}_\text{3D})$ is the expectation of $\text{IoU}_\text{3D}$.}
\resizebox{0.95\textwidth}{!}{
\begin{tabular}{c|ccc|ccc|ccc|ccc|cc}
\toprule[1.2pt]
\multirow{2}{*}{Models} & \multicolumn{3}{c}{SingleBlock} & \multicolumn{3}{c}{TwoBlock} & \multicolumn{3}{c}{ThreeBlock} & \multicolumn{3}{c}{Spatial} & \multicolumn{2}{c}{Total} \\ 
& SR & $\text{IoU}_\text{3D}$ & $e_{trans}$ & SR & $\text{IoU}_\text{3D}$ & $e_{trans}$ & SR & $\text{IoU}_\text{3D}$ & $e_{trans}$ & SR & $\text{IoU}_\text{3D}$ & $e_{trans}$ & SR & $\mathcal{E}(\text{IoU}_\text{3D})$  \\ \hline\hline

Independent & 4/4 & 0.5216 & 1.08 & 4/10 & 0.4608 & 1.88 & 6/10 & 0.4816 & 1.67  & 3/8 & 0.1261 & 3.8    & 17/32 & 0.2605 \\
Mix Blocks & 3/4 & 0.6240 & 0.48 & 3/10 & 0.4378 & 1.94 & 4/10 & 0.5334 & 1.42 & 0/8   & \times & \times & 10/32 & 0.2032 \\
Mix ALL    & 4/4 & 0.6186 & 0.49 & 4/10 & 0.5711 & 0.77 & 2/10 & 0.4056 & 2.11 & 2/8   & 0.5655 & 1.22   & 12/32 & 0.2674 \\

  \bottomrule[0.5pt]
\end{tabular}
}
\label{tab:mix_result}
\end{table*}

%% file: tables/distract.tex
\begin{table*}[htbp]
\centering
\caption{\textbf{Result of baseline models in different distract factors}. $\mathcal{E}(\text{IoU}_\text{3D})$ is the expectation of $\text{IoU}_\text{3D}$.}
\resizebox{0.95\textwidth}{!}{
\begin{tabular}{c|ccc|ccc|ccc|ccc|cc}
\toprule[1.2pt]
\multirow{2}{*}{Models} & \multicolumn{3}{c}{SingleBlock} & \multicolumn{3}{c}{TwoBlock} & \multicolumn{3}{c}{ThreeBlock} & \multicolumn{3}{c}{Spatial} & \multicolumn{2}{c}{Total} \\ 
& SR & $\text{IoU}_\text{3D}$ & $e_{trans}$ & SR & $\text{IoU}_\text{3D}$ & $e_{trans}$ & SR & $\text{IoU}_\text{3D}$ & $e_{trans}$ & SR & $\text{IoU}_\text{3D}$ & $e_{trans}$ & SR & $\mathcal{E}(\text{IoU}_\text{3D})$   \\ \hline\hline
Baseline     & 4/4 & 0.5216 & 1.08   & 4/10 & 0.4608 & 1.88   & 6/10 & 0.4816 & 1.67   & 3/8 & 0.1261 & 3.8    & 17/32 & 0.2605   \\
\hline
Light        & 2/4 & 0.5676 & 0.97   & 0/10 & \times & \times & 2/10 & 0.4365 & 1.95   & 0/8 & \times & \times & 4/32  & 0.0928              \\
Robot State  & 4/4 & 0.5145 & 1.13   & 3/10 & 0.5072 & 1.33   & 5/10 & 0.4584 & 1.74   & 0/8 & \times & \times & 12/32 & 0.2239            \\
Instruction  & 3/4 & 0.3351 & 2.26   & 2/10 & 0.3849 & 2.85   & 0/10 & \times & \times & 0/8 & \times & \times & 5/32  & 0.0821         \\
Background   & 0/4 & \times & \times & 0/10 & \times & \times & 0/10 & \times & \times & 0/8 & \times & \times & 0/32  & 0.0000     \\
Sensor Noise & 2/4 & 0.3677 & 1.93   & 1/10 & 0.3972 & 1.52   & 1/10 & 0.4713 & 1.68   & 0/8 & \times & \times & 4/32  & 0.0677  \\
\hline
Exposure+    & 1/4 & 0.5409 & 1.02 & 0/10 & \times & \times & 1/10 & 0.4411 & 1.92 & 0/8 & \times & \times & 2/32 & 0.0448 \\
Exposure-    & 1/4 & 0.5743 & 0.76 & 0/10 & \times & \times & 1/10 & 0.4027 & 2.36 & 0/8 & \times & \times & 2/32 & 0.0459 \\
  \bottomrule[0.5pt]
\end{tabular}
}
\label{tab:distract_result}
\end{table*}

%% file: sec/5_data.tex
\section{Data Quality and Documentation}

\subsection{Data Collection}

This benchmark is constructed from a curated collection of robot manipulation data covering a diverse set of tasks. Data is collected under clearly defined and standardized protocols to ensure consistency and reproducibility across tasks. Each data sample corresponds to a complete task episode, recorded from task initialization to termination.

Data collection is conducted via teleoperation using the Gello~\citep{wu2024gello} framework. The robotic platform (Franka) is equipped with three depth cameras, including two fixed cameras positioned in front of the robot to capture the task workspace and one wrist-mounted camera attached to the end-effector. In addition to visual observations, all robot state variables are synchronously recorded at each time step throughout data collection. Demonstrations are generated by trained human operators following task-specific guidelines. To capture natural variability, task instances are executed under different initial conditions, object configurations, and environmental states. All data collection procedures are applied uniformly across tasks to minimize uncontrolled variation.

\subsection{Data Organization and Structure}

We adopt the \texttt{LeRobotDataset} format~\citep{cadene2024lerobot} to package our real-robot demonstrations into a unified representation. Following LeRobot's convention, the dataset is organized as a repository containing a \texttt{data/} folder for per-episode trajectories stored in Parquet files, and a \texttt{meta/} folder that centralizes the dataset schema, indexing, task annotations, and statistics. Concretely, each episode is stored as a time-series table with synchronized multi-modal observations and actions.

\paragraph{Modalities and temporal resolution.}
Each timestep includes three RGB image streams and low-dimensional robot state/action vectors:
\begin{itemize}
    \item \textbf{Images:} \texttt{image}, \texttt{wrist\_image}, \texttt{second\_image}, all stored as raw RGB frames of size $640 \times 480$.
    \item \textbf{Proprioception/state:} \texttt{state} $\in \mathbb{R}^{8}$ (float32).
    \item \textbf{Actions:} \texttt{actions} $\in \mathbb{R}^{8}$ (float32).
    \item \textbf{Indexing:} \texttt{timestamp}, \texttt{frame\_index}, \texttt{episode\_index}, and global \texttt{index}; plus \texttt{task\_index} for language-conditioned supervision.
\end{itemize}

We keep the original image resolution and leave resizing/cropping and other vision augmentations to the model-side preprocessing pipeline. This design improves compatibility with diverse VLA backbones that may require different image sizes and camera preprocessing policies.

\subsection{Metadata and Documentation}

LeRobot uses a set of metadata files under \texttt{meta/} to make datasets self-describing and easy to index/search/visualize. In our release, the following files are provided:

\paragraph{\texttt{meta/info.json} (dataset schema \& paths).}
This is the central metadata file: it defines the complete feature schema (dtype/shape), FPS, codebase version, and path templates that locate data files. Such a schema-centric design enables downstream code to validate shapes/dtypes and to programmatically enumerate episodes without any external assumptions.

\paragraph{\texttt{meta/tasks.jsonl} (task inventory).}
It enumerates all language instructions (task strings) and maps each task to a unique integer \texttt{task\_index}. This makes the dataset language-conditioned while allowing efficient filtering and stratified sampling by task.

\paragraph{\texttt{meta/episodes.jsonl} (per-episode annotations).}
It stores per-episode metadata such as:
\begin{itemize}
    \item \texttt{episode\_index}: integer episode id.
    \item \texttt{tasks}: a list of language instructions aligned with the episode (in our case typically one instruction per episode, e.g., ``Pick up the red block and put in the target place.'').
    \item \texttt{length}: number of timesteps/frames in the episode (variable across episodes).
\end{itemize}
This file provides the primary index for dataset browsing and for sampling episodes by length/task.

\paragraph{\texttt{meta/episodes\_stats.jsonl} (per-episode statistics).}
It contains per-episode summary statistics (e.g., min/max/mean) for selected modalities, which is useful for dataset sanity checks and debugging (e.g., detecting corrupted frames, constant signals, or abnormal ranges). LeRobot explicitly supports storing such statistics as part of the dataset metadata.

\paragraph{\texttt{norm\_stats.json} (dataset-level normalization statistics).}
Following \cite{black2024pi_0,intelligence2504pi0} data processing protocol, after dataset conversion, we run a dedicated script to scan the entire dataset and compute global statistics for low-dimensional signals (state and actions). Specifically, we aggregate all frames across all episodes to obtain the per-dimension mean and standard deviation, together with robust percentiles (q01, q99). The resulting statistics are saved and are used to standardize or robustly clip inputs consistently across training runs and model variants.

\begin{figure*}[t]
    \centering
    \includegraphics[width=1.\textwidth]{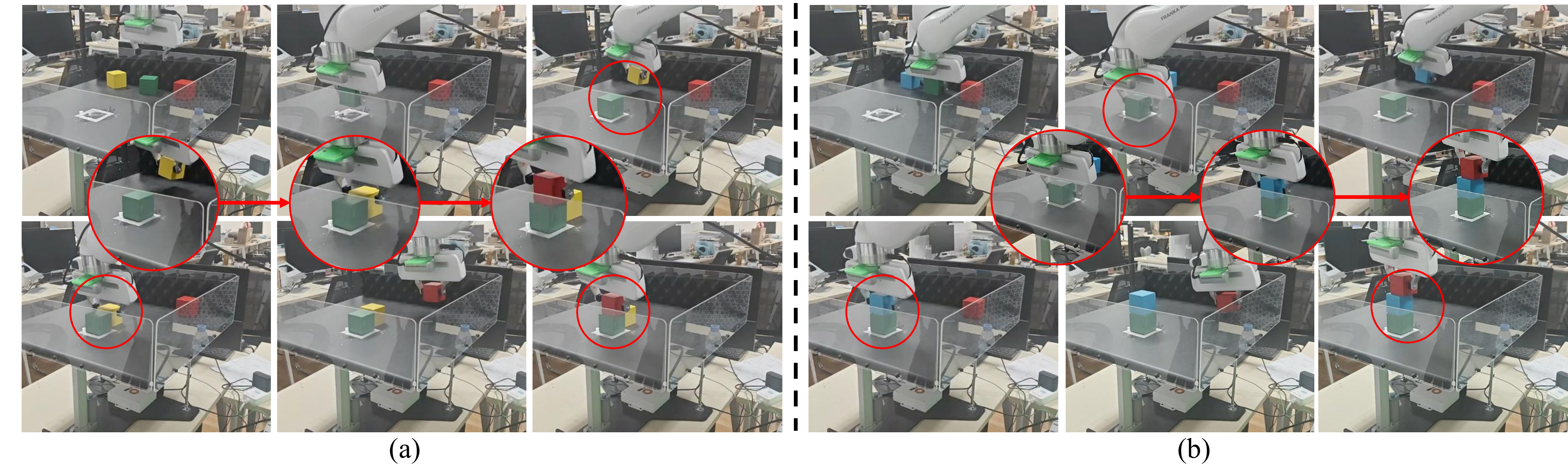}
    \caption{\small{\textbf{Visualization of two cases of ThreeBlocks split.} 
}}
    \label{fig:three}
\end{figure*}

\begin{figure*}[t]
    \centering
    \includegraphics[width=1.\textwidth]{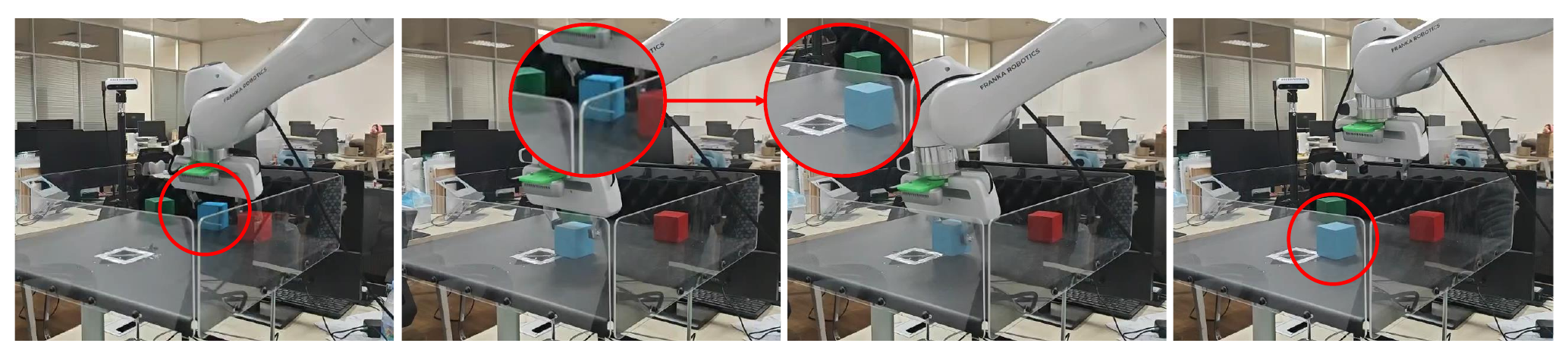}
    \caption{\small{\textbf{Visualization of case of Spatial split.} 
}}
    \label{fig:spatial}
\end{figure*}

\begin{figure*}[t]
    \centering
    \includegraphics[width=1.\textwidth]{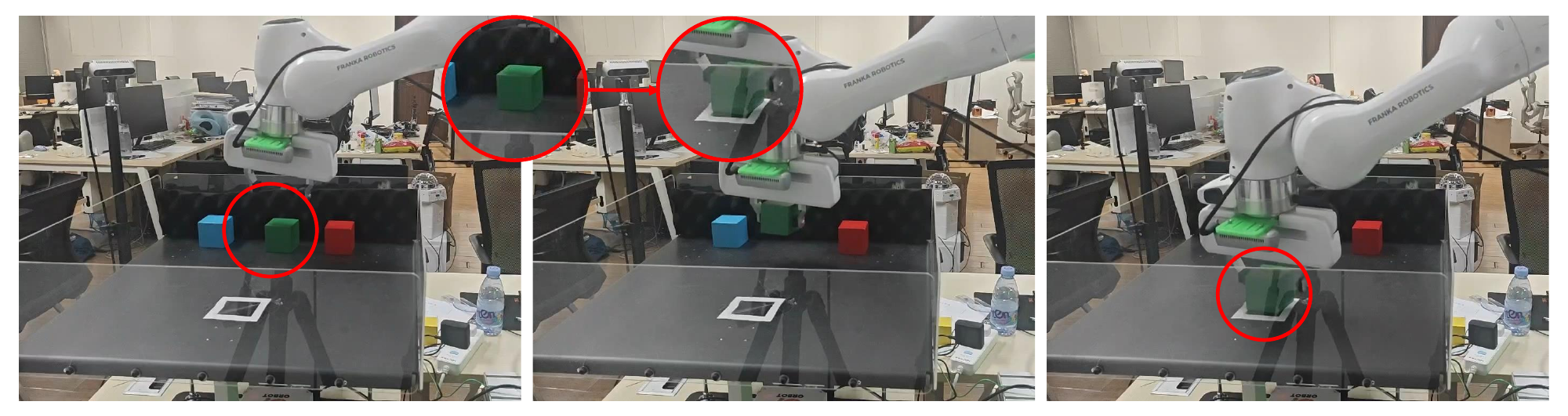}
    \caption{\small{\textbf{Visualization of case of SingleBlock split.} 
}}
    \label{fig:one}
\end{figure*}

\begin{figure*}[t]
    \centering
    \includegraphics[width=1.\textwidth]{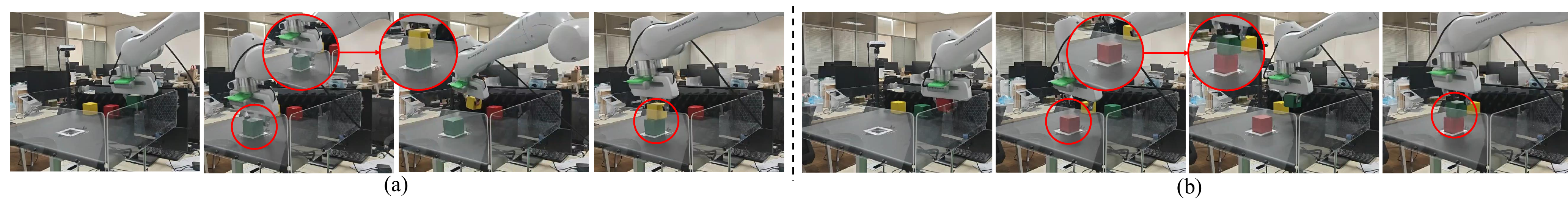}
    \caption{\small{\textbf{Visualization of two cases of TwoBlocks split.} 
}}
    \label{fig:two}
\end{figure*}

\begin{figure*}[t]
    \centering
    \includegraphics[width=1.\textwidth]{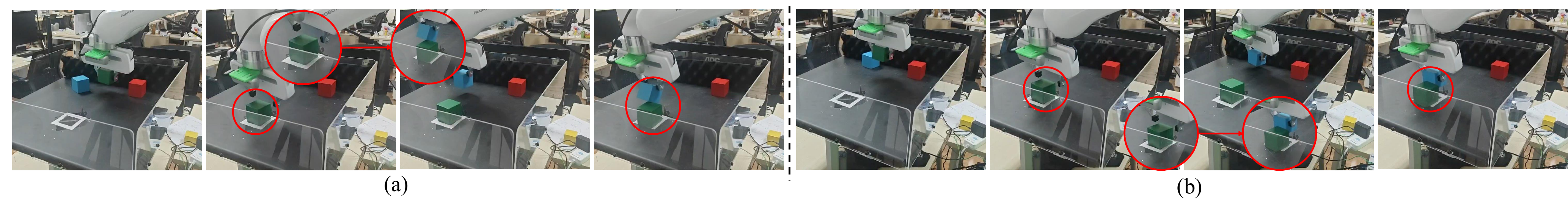}
    \caption{\small{\textbf{Visualisation of two failed cases.} 
}}
    \label{fig:fail}
\end{figure*}